\documentclass[review=true,anonymous,runningheads]{llncs}

\usepackage{hyperref}

\usepackage{mathrsfs}
\usepackage{mathtools}
\usepackage{acronym}
\usepackage{booktabs}
\usepackage[skip=0pt]{caption}
\usepackage[capitalise]{cleveref}
\usepackage[T1]{fontenc}

\usepackage{graphicx}
\usepackage{listings}
\usepackage{multirow}
\usepackage{subcaption}
\usepackage{tabularx}
\usepackage[dvipsnames]{xcolor}
\usepackage{colortbl}
\usepackage{calc}
\usepackage[inline]{enumitem}
\usepackage{numprint}
\usepackage[normalem]{ulem}

\usepackage{balance}
\usepackage{xspace}

\usepackage{tikz}
\usetikzlibrary{positioning, fadings,patterns,fit}
\usetikzlibrary{arrows}

\definecolor{ali}{RGB}{0, 150, 0}

\definecolor{Gray}{gray}{0.8}
\definecolor{Gray2}{gray}{0.93}

\newcommand{\myparagraph}[1]{\vspace*{0.7mm}\emph{\bfseries #1}}

\newtheorem{observation}{Observation}

\newcommand{\llmb}{LLM-1}
\newcommand{\llmm}{LLM-2}
\newcommand{\llmo}{LLM-3}

\author{Ali Vardasbi \and
Gustavo Penha \and
Claudia Hauff \and
Hugues Bouchard
}
\authorrunning{A. Vardasbi et al.}
\institute{Spotify \\
\email{\{aliv,gustavop,claudiah,hb\}@spotify.com}}

\allowdisplaybreaks

\begin{document}

\title{Adaptive Repetition for Mitigating Position Bias in LLM-Based Ranking}

\maketitle

\begin{abstract}
When using LLMs to rank items based on given criteria, or evaluate answers, the order of candidate items can influence the model's final decision.
This sensitivity to item positioning in a LLM's prompt is known as \emph{position bias}.
Prior research shows that this bias exists even in large models, though its severity varies across models and tasks.
In addition to position bias, LLMs also exhibit varying degrees of low \emph{repetition consistency}, where repeating the LLM call with the same candidate ordering can lead to different rankings.  
To address both inconsistencies, a common approach is to prompt the model multiple times with different candidate orderings and aggregate the results via majority voting.  
However, this repetition strategy, significantly increases computational costs.  

Extending prior findings, we observe that both the direction---favoring either the earlier or later candidate in the prompt---and magnitude of position bias across instances vary substantially, even within a single dataset.  
This observation highlights the need for a \emph{per-instance} mitigation strategy.  
To this end, we introduce a dynamic early-stopping method that adaptively determines the number of repetitions required for each instance.
Evaluating our approach across three LLMs of varying sizes and on two tasks, namely re-ranking and alignment, we demonstrate that transitioning to a dynamic repetition strategy reduces the number of LLM calls by an average of \( 81\% \), while preserving the accuracy.  
Furthermore, we propose a confidence-based adaptation to our early-stopping method, reducing LLM calls by an average of \(87\%\) compared to static repetition, with only a slight accuracy trade-off relative to our original early-stopping method.


\end{abstract}



\section{Introduction}

LLMs are increasingly used to select a winner from a set of items in various tasks, such as comparing the responses of different LLMs to questions \cite{wang2024largelanguagemodels,shi2024judgingthejudges,li2024fromgenerationto}, re-ranking documents based on their relevance to a query \cite{qin2024largelanguagemodels}, or answering multiple-choice questions \cite{robinson2024leveraginglargelanguage,zheng2024largelanguagemodels}.
Here, we focus on scenarios where an LLM is given a set of items and asked to select \emph{one} as the top-ranked item or winner. 
Accordingly, we use the terms \emph{ranking} and \emph{judgment} interchangeably throughout the paper.
Previous research has shown that LLMs, even highly capable models, exhibit position bias \cite{shi2024judgingthejudges} \cite{ye2024justiceorprejudice?}, which refers to the inconsistency in the LLM’s verdict when the order of candidates within one prompt is changed. We refer to this as a lack of \emph{permutation consistency}~(PC).
Additionally, \emph{repetition consistency} (RC) measures the stability of the LLM’s responses when the same prompt is repeated with the same ordering. A low RC suggests that the model may be uncertain or sensitive to minor variations.
Together, high values of PC and RC could be indicators of a \emph{low-variance} judgment, as they reflect the stability and consistency of the model’s decisions.

Previous studies \cite{shi2024judgingthejudges,wang2024largelanguagemodels,chen2024mllmasajudgeassessingmultimodalllmasajudge} have shown that repeating judgments with different permutations and selecting the majority outcome enhances robustness.  
Our experiments confirm these findings, demonstrating that increasing the number of repetitions\footnote{We observe a plateau in accuracy beyond \( 24 \) repetitions.} reduces variance and improves average accuracy.  
However, more repetitions also lead to higher computational costs.  
\emph{Ideally, we are interested in maintaining similar accuracy levels while minimizing cost.}



Our contributions in this paper are as follows:
\begin{enumerate*}[label={\textbf{(\arabic*)}}]
    \item Our experiments on five different datasets reveal that the position bias of LLMs is not only task-dependent but can also vary across instances. 
    Specifically, while an LLM may show a preference for earlier-positioned candidates in one judgment instance, it may exhibit a preference for later-positioned candidates in another instance within the same dataset (\cref{sec:results:preference}).
    This highlights the need for a per-instance treatment of position bias in LLMs.
    \item We propose an effective approach to address position bias on a per-instance basis: adaptively determining the number of required repetitions for each instance, significantly reducing the number of LLM calls by \( 81\%\) on average (hence lowering the cost) while maintaining accuracy at a comparable level.
    Our early-stopping method (\cref{sec:judgement:earlystopping}) strikes a balance between accuracy and efficiency. For harder instances, where LLM judgments are less stable, we perform additional repetitions to improve reliability. However, on average, the total number of LLM calls remains low, ensuring computational efficiency without sacrificing ranking robustness—achieving the best of both worlds.
    \item We further refine our early-stopping method by estimating repeat inconsistency based on the LLM's confidence (\cref{sec:judgement:confidencebased}), reducing LLM calls by an average of \(87\%\) compared to the static repetition strategy.
\end{enumerate*}

\myparagraph{Disclaimer.}
This paper is not focused on comparing LLMs or evaluating their relative performance; therefore, we do not name specific models.
Our goal is to highlight the presence of position bias and repetition inconsistency, and to demonstrate that these issues can be effectively mitigated.
We observed consistent behaviors across three LLMs of varying sizes (from moderately sized to state-of-the-art large-scale LLMs), including both proprietary and open-source models, and report results from these representative models to illustrate the effectiveness of our approach.
We refer to these LLM models simply by \llmb{}, \llmm{}, and \llmo{}.


\section{Robust Rankings}
\label{sec:judgement}

\begin{figure}[t]
    \centering
    \includegraphics[width=0.6\linewidth]{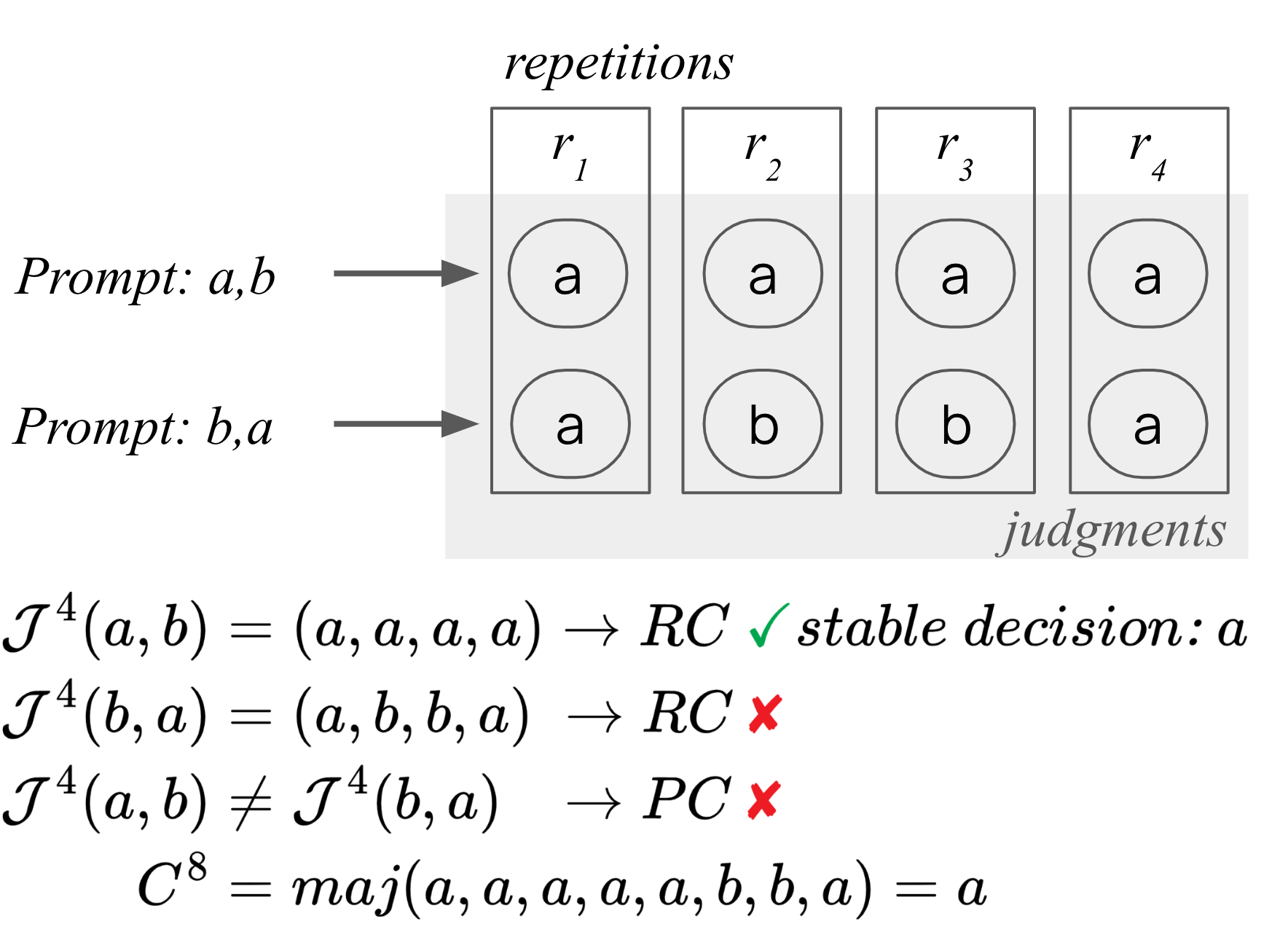}
    \caption{Example outcomes and corresponding RC, PC and $C^{2n}$ values.}
    \label{fig:example}
\end{figure}

Assume an LLM-based ranking scenario, where an LLM is used for pairwise comparisons over a list of paired candidates \(\{(a_i, b_i)\}_{i=1}^{N}\).
We denote the outcome of the LLM when prompted to pick one from the \emph{ordered} candidates \((a, b)\) as \(J(a, b)\). 
Repeating the ranking task \(n\) times with the same ordering of candidates results in a vector of \(n\) outcomes, which we denote as \(\mathcal{J}^n(a, b) \in \{a,b\}^n\). See~\cref{fig:example} for a concrete example.

\myparagraph{Repetition Consistency.}
We consider the LLM for the ordered candidates \( (a , b) \) to be repetition consistent (RC) after \( n \) repetitions if the \( \mathcal{J}^n(a, b) \) vector consists of a unique verdict, and we call the unique verdict the \emph{stable decision}.

\myparagraph{Permutation Consistency.}
We consider an LLM to be permutation consistent (PC) for candidates \( (a , b) \) after \( n \) repetitions if all outcomes for both input orderings are identical. 
In other words, \( (a, b) \) and \( (b, a) \) are both repetition consistent (RC), and the vectors \( \mathcal{J}^n(a, b) \) and \( \mathcal{J}^n(b, a) \) contain the \emph{same} stable decisions.

It is worth noting that the definitions of RC and PC depend on both the specific experiment and the number of repetitions \( n \). Throughout this paper, references to RC and PC assume a finite, practical number of repetitions. Similarly, when we refer to stable rankings, we mean stability within the context of the given experiment and repetition count; this does not necessarily imply stability for larger numbers of repetitions.

\myparagraph{Consensus Outcome.}
The consensus outcome after \( 2n \) repetitions, denoted as \( C^{2n}(a, b) \in \{a, b, \textit{tie}\} \), is determined by majority voting over the concatenation of \( \mathcal{J}^n(a, b) \) and \( \mathcal{J}^n(b, a) \).
If \( a \) and \( b \) appear an equal number of times in the concatenated vector, the consensus outcome is deemed \emph{inconclusive}, resulting in a \emph{tie}.\looseness=-1

As discussed earlier, prior studies show that increasing \( n \) improves the accuracy of consensus outcome \cite{chen2024mllmasajudgeassessingmultimodalllmasajudge}.
However, the efficiency decreases as \( n \) grows.  
Next, we propose two early-stopping methods that significantly reduce the average number of LLM repetitions while preserving the accuracy of the consensus outcome.

\myparagraph{Remark.}
In this paper, our experiments focus exclusively on pairwise rankings. Our findings can be generalized to listwise rankings by replacing the swapping of two items with different permutations of the candidate list. While this introduces exponential complexity as the number of items increases, prior work (e.g., \cite{zheng2024largelanguagemodels}) has shown that \emph{cyclic permutations} offer a practical alternative with linear complexity.

\myparagraph{Observation.}
We first present a key observation that serves as the foundation of our early-stopping strategies in the following sections. 
Across extensive experiments, we found that for nearly all candidate pairs, the LLM's outcome after multiple repetitions remains consistent for at least one input ordering.
We hypothesize that this observation occurs when the judgment of the LLM regarding the winner candidate coincides with its position bias for that pair of candidates (see \cref{sec:results:preference}).
To elaborate, if an LLM exhibits a positional preference for the earlier candidate and also inherently favors candidate $a$ over $b$ based on its reasoning, then when $a$ is placed in the first position, the model's outcome remains consistent.  
In such cases, the LLM's inherent preference aligns with the positional bias, reinforcing the selection of $a$ and leading to a stable decision.
Conversely, when the LLM's intrinsic preference contradicts its position bias for a given order, inconsistency may arise across different repetitions of its judgment.
Formally:

\begin{observation}
\label{observation:rc}
    With high empirical probability, for each LLM and candidate pair \( (a, b) \), the LLM is RC for at least one of the orderings \( (a, b) \) or \( (b, a) \).  
    This implies that, with only a small fraction of exceptions, the LLM outcome for one of the orderings remains consistent as the number of repetitions increases.
\end{observation}

\begin{table}[t]
\small
    \centering
    \caption{Violation rates for \cref{observation:rc} with \( n=12 \).}
    \label{tab:observation}
\begin{tabular}{l c c c}
\toprule
 Dataset & \cellcolor[gray]{0.75} \llmb & \cellcolor[gray]{0.92} \llmm & \llmo \\
\midrule
MSMarco & \cellcolor[gray]{0.75} 0.97\% & \cellcolor[gray]{0.92} 0.69\% & 4.5\% \\
Emerton-DPO & \cellcolor[gray]{0.75} 6.3\% & \cellcolor[gray]{0.92} 3.1\% & 1.4\% \\
Orca-DPO & \cellcolor[gray]{0.75} 5.5\% & \cellcolor[gray]{0.92} 3.5\% & 0.87\% \\
Py-DPO & \cellcolor[gray]{0.75} 2.6\% & \cellcolor[gray]{0.92} 4.8\% & 0.7\% \\
Truthy-DPO & \cellcolor[gray]{0.75} 1.7\% & \cellcolor[gray]{0.92} 1.5\% & 0.76\% \\
\bottomrule
\end{tabular}
\end{table}

\Cref{tab:observation} shows the violation rates of this observation for different LLMs and datasets (see \cref{sec:experiments}, Datasets).
Here, the violation rate means the percentage of the candidate pairs for which none of the orderings are RC for \(n=12\).
As is seen in this table, the this percentage remains small across models and datasets. 

Based on \cref{observation:rc}, we propose two early stopping approaches to improve LLM robustness while minimizing computational costs.  
We validate our methods across three LLMs and five diverse datasets (see \cref{sec:experiments}).

\subsection{Early Stopping}
\label{sec:judgement:earlystopping}
Our early stopping criterion is an approximation that assumes \cref{observation:rc} holds with probability \( 1 \).  
In other words, for the small fraction of violating instances reported in \cref{tab:observation}, our early stopping criterion may not yield the consensus outcome.  
However, for the vast majority of instances, it is guaranteed to converge to the consensus outcome.

Our early stopping works as follows.
Starting from \( n=1 \), we prompt the LLM-ranker with both orderings \( (a, b) \) and \( (b, a)\) and monitor \( C^{2n}(a, b) \).
We stop the repetition as soon as we reach a conclusive majority voting outcome.
In other words, we increment \( n \) \emph{only} if the majority voting is inconclusive, i.e., when \( C^{2n}(a, b) = \textit{tie}\).

In what follows we discuss why this method leads to consensus outcome.
For each \( n \), one of the following situations must occur:
\begin{enumerate*}[label=(\Roman*)]
    \item Both orderings are RC but with different stable decisions, e.g. \( \mathcal{J}^3(a, b)=(a,a,a) \) and \( \mathcal{J}^3(b, a)=(b,b,b) \).
    \label{sec:judgment:earlystopping:aabb}
    \item Both orderings are RC with the same stable decisions.
    \label{sec:judgment:earlystopping:aaaa}
    \item Only one of the vectors is RC, while the other contains mixed outcomes.
    \label{sec:judgment:earlystopping:aaba}
    \item Neither of the vectors is RC.
    \label{sec:judgment:earlystopping:abab}
\end{enumerate*}

First, note that the last case, i.e., \ref{sec:judgment:earlystopping:abab}, occurs with low empirical probability according to \cref{tab:observation}. 
Therefore, we ignore this case, as its rarity means it contributes negligibly to the overall accuracy (as validated experimentally in \cref{sec:results:performance}).
Among the remaining cases, only in case \ref{sec:judgment:earlystopping:aabb}, where each RC vector has a different stable decision, will \( C^{2n}(a, b) \) be inconclusive, requiring additional repetitions. 
In the other two cases, without loss of generality, assume that the stable decision of the RC vector is \( a \). 
Since the other vector is not RC with \( b \), it must contain at least one outcome that differs from \( b \) (either \( a \) or producing a \emph{tie}).  
As a result, the majority vote in the concatenated vector will favor \( a \), and this consensus will remain unchanged (with high probability) as \( n \) increases because, according to \cref{observation:rc}, the ordering will continue to be RC and consistently generate \( a \).
Thus, by detecting the RC ordering early and stopping as soon as a conclusive majority is reached, we significantly reduce the number of repetitions required while preserving decision accuracy.

\myparagraph{Example.}  
Suppose the outcome vectors for a given instance are \( \mathcal{J}^2(a, b)=(a, a) \) and \( \mathcal{J}^2(b, a)=(b, b) \).
For \( n=1 \) and \( n=2 \), we fall into case \ref{sec:judgment:earlystopping:aabb}: it remains uncertain which ordering will be RC for larger \( n \).
However, at \( n=3 \), we observe that only the first vector is RC: \( \mathcal{J}^3(a, b)=(a, a, a) \) and \( \mathcal{J}^3(b, a)=(b, b, a) \) (case \ref{sec:judgment:earlystopping:aaba}).
Note that according to \cref{observation:rc}, since the ordering \( (b, a) \) is not RC, with high probability the other ordering remains RC even for larger values of \( n \), i.e., the ordering \( (a, b) \) will consistently generate \( a \).
Consequently, the consensus outcome stabilizes at \( a \) for all \( n \geq 3 \).

\subsection{Confidence-Based Early Stopping}
\label{sec:judgement:confidencebased}
In the early stopping method discussed in \cref{sec:judgement:earlystopping}, when two contradicting RC vectors occur, i.e., case \ref{sec:judgment:earlystopping:aabb}, the repetition must continue until a tie is broken.  
The challenge arises when the vectors remain contradictory even after reaching the maximum number of repetitions.  
If we could predict this outcome in advance, we could declare such instances a \emph{tie} sooner.

To achieve this, let's assume we can estimate the probability gap \( g = | P_a - P_b | \) between the two candidates based on a single sample of the paired judgment results: \( \big( J(a, b), J(b, a) \big) \), without directly knowing which candidate is more probable (since otherwise, no repetitions would be necessary). 
Let \( n_M \) denote the maximum number of paired repetitions.
If our estimated gap \( g \) is accurate, and we were to repeat the judgment for each ordering \( n_M \) times, then for sufficiently large \( n_M \), we would observe approximately \( (1-g)n_M \) and \( (1+g)n_M \) of the minority and majority outcomes, respectively, in the concatenated vector of paired repeated judgments.
In the worst-case scenario, the first \( (1-g)n_M \) paired judgments could fall into case \ref{sec:judgment:earlystopping:aabb} (as described in \cref{sec:judgement:earlystopping}), leading to an inconclusive outcome. Therefore, at most \( (1-g)n_M + 1 \) paired repetitions are needed to guarantee a conclusive result.

For now, assume we have access to an oracle that provides \( g \) for a given candidate pair \( (a, b) \) and the LLM ranker. Our approach is as follows:
We adapt the early stopping method in \cref{sec:judgement:earlystopping} by limiting the number of paired repetitions to \( (1-g)n_M + 1 \), instead of the full \( n_M \), where \( g \) is specific to the candidate pair instance.

\myparagraph{Confidence Gap.}
Now, we need to define the method that can be used to estimate the probability gap \( g =| P_a - P_b | \). 
For this, we prompt the LLM judges to include a \emph{confidence} value in their judgments, indicating their level of confidence in the generated answer \cite{xiong2024llmsexpressuncertaintyempirical}. Our experiments show a correlation between the confidence gap — i.e., the gap between the average generated confidence values of the LLM when outputting \( a \) compared to \( b \) — and the probability gap \( g \). 
As such, we use the confidence gap to estimate the probability gap. 
To make the estimate data-driven, we sample a small percentage of instances (\( 10\% \)) to fit a linear model that predicts the probability gap from the confidence gap.
When reporting the average number of calls (\cref{tab:runs}), we consider the number of calls for this training data to be \( 2\times n_M\) (24 in our tests).

\myparagraph{Remark.}
Our notion of probability gap is inspired by prior research showing that the quality gap between candidates can influence the magnitude of position bias \cite{shi2024judgingthejudges} \cite{ye2024justiceorprejudice?} \cite{wang2024largelanguagemodels}.

\section{Experimental Setup}
\label{sec:experiments}

\myparagraph{Datasets.}
We have used the following datasets in our LLM-as-a-judge experiments:
\begin{itemize}
    \item {\textbf{Re-rank:}} We used the test sets from the 2019 and 2020 TREC Deep Learning (TREC-DL) competitions: TREC-DL 2019 and TREC-DL 2020, both of which provide dense human relevance annotations for each query. These datasets are based on the MS MARCO v1 document corpus.
    In total, there are 86 unique queries, with 1,449 sampled document pairs. For each query, we sampled documents from the top 100 retrieved by the original dataset provider. Additionally, we ensured that each sampled document pair had a minimum relevance score gap of 2, as per the provided annotations.
    \item {\textbf{Alignment:}} Following \cite{ye2024justiceorprejudice?}, we use four DPO (Direct Preference Optimization) datasets \cite{rafailov2024directpreferenceoptimizationlanguage}, where each dataset is composed of user feedback collected across various scenarios. Each prompt in these datasets is paired with a positive (``chose'') and a negative (``reject'') answer, reflecting user preferences and rejections. The datasets include:
    \begin{enumerate*}[label=({\Roman*})]
    \item Emerton-DPO \cite{emertondpo} with 839 samples,
    \item Orca-DPO \cite{orcadpo} with 923 samples,
    \item Py-DPO \cite{pydpo} with 1003 samples, and
    \item Truthy-DPO \cite{truthydpo} with 654 samples.
    \end{enumerate*}
\end{itemize}

\myparagraph{Models.}
We run our experiments on three LLMs of varying capabilities and sizes: an open-source and two proprietary models of varying scales, from mid-sized to some of the largest publicly or commercially available LLMs. We refer these by \llmb{}, \llmm{}, and \llmo{}.

\textit{\myparagraph{Remark.} This study does not aim to compare these LLMs against each other. Instead, it seeks to demonstrate that repeated LLM calls can enhance ranking consistency across a spectrum of models, from compact to large-scale. The selected models are intentionally diverse in capacity to illustrate this effect comprehensively.}\looseness=-1

\myparagraph{Parameters.}
We set the maximum number of repetitions for the consensus judgment to \(24\) (\(n_M = 12\)). While RC and PC values may slightly change with a higher repetition cap, we consider \(24\) repetitions to be beyond the practical threshold for LLM-based judgments. Moreover, our experiments show that overall judgment accuracy plateaus after this point, indicating diminishing returns from additional repetitions.
For all LLM calls, we set the temperature to \( 0.1 \) to reduce variance in the model’s responses while still allowing some degrees of diversity. This choice aligns with prior work \cite{shi2024judgingthejudges,ye2024justiceorprejudice?}, which similarly adopted non-zero temperature values to avoid trivial or overly deterministic outputs.


\section{Results}
In this section, we first demonstrate that position bias (indicated by imperfect RC and PC values) in LLMs not only varies in magnitude but can also vary in direction within a single dataset. 
This leads to LLM preferences for earlier items in some judgment instances and for later items in others. 
Consequently, attempting to address position bias with a single treatment for each task results in suboptimal performance. 
This suggests that the most robust approach to mitigating a lack of RC and PC is to conduct a large number of paired repetitions and select the consensus judgment.

Subsequently, we present the cost reduction results obtained from our early-stopping methods in \cref{sec:judgement}, demonstrating how these reductions are achieved while preserving the robustness and accuracy of the consensus judgment.

\begin{table*}[h]
    \centering
    \small
    \caption{Comparison of bias metrics for different LLMs on different datasets.}
    \label{tab:bias-llm}
    \begin{tabular}{l c c c c c c c c c}
    \toprule
    \multirow{2}{*}{Dataset} & \multicolumn{3}{c}{PC} & \multicolumn{3}{c}{Primacy Biased} & \multicolumn{3}{c}{Recency Biased} \\
    \cmidrule(lr){2-4} \cmidrule(lr){5-7} \cmidrule(lr){8-10}

     & \cellcolor[gray]{0.75} \scriptsize \llmb{} & \cellcolor[gray]{0.92} \scriptsize \llmm{} & \scriptsize \llmo{} & \cellcolor[gray]{0.75} \scriptsize \llmb{} & \cellcolor[gray]{0.92} \scriptsize \llmm{} & \scriptsize \llmo{} & \cellcolor[gray]{0.75} \scriptsize \llmb{} & \cellcolor[gray]{0.92} \scriptsize \llmm{} & \scriptsize \llmo{} \\
    \midrule

    \scriptsize MSMarco & \cellcolor[gray]{0.75} 0.726 & \cellcolor[gray]{0.92} 0.798 & 0.436 & \cellcolor[gray]{0.75} 0.030 & \cellcolor[gray]{0.92} 0.152 & 0.334 & \cellcolor[gray]{0.75} 0.244 & \cellcolor[gray]{0.92} 0.050 & 0.230 \\ 
    \scriptsize Emerton-DPO & \cellcolor[gray]{0.75} 0.371 & \cellcolor[gray]{0.92} 0.308 & 0.239 & \cellcolor[gray]{0.75} 0.303 & \cellcolor[gray]{0.92} 0.263 & 0.632 & \cellcolor[gray]{0.75} 0.326 & \cellcolor[gray]{0.92} 0.429 & 0.129 \\ 
    \scriptsize Orca-DPO & \cellcolor[gray]{0.75} 0.547 & \cellcolor[gray]{0.92} 0.462 & 0.325 & \cellcolor[gray]{0.75} 0.295 & \cellcolor[gray]{0.92} 0.247 & 0.552 & \cellcolor[gray]{0.75} 0.158 & \cellcolor[gray]{0.92} 0.292 & 0.123 \\ 
    \scriptsize Py-DPO & \cellcolor[gray]{0.75} 0.694 & \cellcolor[gray]{0.92} 0.557 & 0.313 & \cellcolor[gray]{0.75} 0.219 & \cellcolor[gray]{0.92} 0.229 & 0.650 & \cellcolor[gray]{0.75} 0.086 & \cellcolor[gray]{0.92} 0.214 & 0.037 \\ 
    \scriptsize Truthy-DPO & \cellcolor[gray]{0.75} 0.787 & \cellcolor[gray]{0.92} 0.673 & 0.680 & \cellcolor[gray]{0.75} 0.188 & \cellcolor[gray]{0.92} 0.120 & 0.222 & \cellcolor[gray]{0.75} 0.025 & \cellcolor[gray]{0.92} 0.208 & 0.099 \\

    \bottomrule

    \end{tabular}
\end{table*}

\subsection{Preference Variation}
\label{sec:results:preference}

First, focusing on PC and RC parameters, we show that the bias of an LLM varies in direction for different judgment instances from the same dataset.
We distinguish between three types of behaviors:
\begin{enumerate*}
\item {\textbf{PC:}} The LLM shows no preference for the position of candidates, demonstrating PC.
\item {\textbf{Primacy Biased:}} The LLM favors the first item in a pair in its prompt.
\item {\textbf{Recency Biased:}} The LLM favors the last item in a pair in its prompt.\looseness=-1
\end{enumerate*} \cref{tab:bias-llm} shows the ratio of instances that fall within each category of behavior. 

While we categorize position bias into primacy and recency directions based on empirical patterns in model judgments, we refrain to claim a singular underlying cause. Position bias may arise from various architectural or representational factors, such as the model's attention dynamics, tokenization artifacts, or prompt formatting, that are difficult to disentangle, especially across different LLMs. Rather than assuming a fixed mechanism, our approach acknowledges the empirical presence of position bias and proposes a generalizable, instance-adaptive method to mitigate its effects. By observing that the direction and magnitude of bias vary across instances, we focus on robustness at inference time through dynamic repetition strategies, without relying on assumptions about model internals. A deeper mechanistic understanding remains an important direction for future work, especially for developing complementary mitigation strategies.


Among the datasets in \cref{tab:bias-llm}, we observe that \llmb{} on Emerton-DPO and \llmm{} on both Orca-DPO and Py-DPO datasets exhibit nearly equal ratios of Primacy and Recency. 
Consequently, averaging the bias direction across all instances of these model-dataset pairs would misleadingly suggest that the models exhibit negligible position bias on these datasets. 
However, noticing the low Consistent ratios, we can see that this is not the case. 
These datasets would benefit the most from a per-instance bias treatment approach.
On the other end of the spectrum, some model-dataset pairs exhibit a noticeably stronger bias in one direction. 
For example, \llmo{} shows a strong Primacy preference across all the datasets, while \llmb{} and \llmm{} display strong Recency and Primacy preferences over the MSMarco dataset, respectively. 
In these cases, a per-instance treatment for position bias, though still important, is less critical than in other scenarios.

\begin{figure}[t]
    \centering
    \begin{tabular}[]{
    @{}l@{\hspace{0.2em}}c@{\hspace{0.2em}}c@{\hspace{0.2em}}c
    }
    
    \multicolumn{4}{c}{\includegraphics[width=0.52\textwidth]{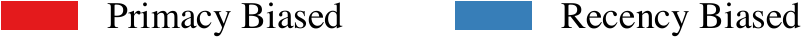}}
    \\
        & Model: \llmb{}
        & Model: \llmm{}
        & Model: \llmo{}
        \\
        \rotatebox[origin=lt]{90}{\scriptsize  \hspace{3.3em} Density}
        & \includegraphics[width=0.3\textwidth]{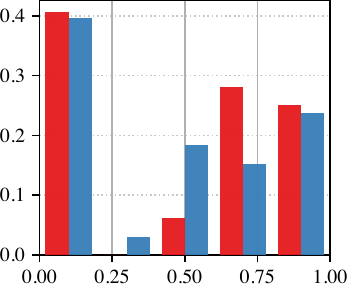}
        & \includegraphics[width=0.3\textwidth]{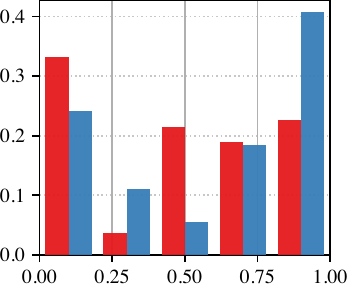}
        & \includegraphics[width=0.3\textwidth]{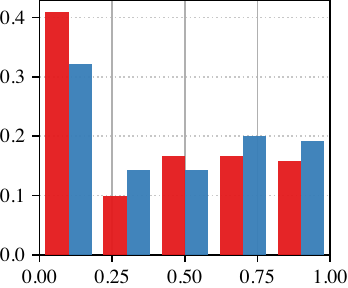}
        \\
        & \scriptsize \hspace{0.8em} Probability Gap
        & \scriptsize \hspace{0.8em} Probability Gap
        & \scriptsize \hspace{0.8em} Probability Gap \vspace{0.7em}
        
    \end{tabular}
    \caption{Histogram of the probability gap \( g =| P_a - P_b | \) on the MSMarco dataset for different LLMs. A probability gap of \( 0 \) with ``Primacy Biased'' means that the first candidate in the input is always the outcome. Higher values for probability gap means lower position bias. }
    \label{fig:probgap}
\end{figure}

\cref{fig:probgap} provides a deeper view into the distribution of probability gaps for candidate pairs in the MSMarco dataset across different models. Notably, a substantial portion of both Primacy and Recency biased pairs fall within the bracket \( [0,0.2) \), indicating a strong position bias in many individual judgments. Crucially, these biases occur in both directions, highlighting that the direction of position bias is not the same across instances of the same dataset. This reinforces our earlier claim: dataset-level averages can obscure substantial instance-level variation, and a single, static mitigation strategy is unlikely to be effective. Instead, a dynamic, per-instance approach is essential for robust judgment.

\cref{fig:probgap} also highlights why our confidence-based early stopping method (\cref{sec:judgement:confidencebased}) achieves better efficiency compared to the standard early stopping approach (\cref{sec:judgement:earlystopping}). Recall that in the confidence-based method, the number of paired repetitions for each instance is bounded by \( (1-g)n_M + 1\), where \( g \) is the estimated probability gap between the two candidates. As shown in the figure, a noticeable number of candidate pairs have large probability gaps, both in the primacy and recency directions. For these high-gap instances, the stopping criterion triggers earlier, reducing the number of required LLM calls.

\subsection{Efficient Robust Bias Treatment}
\label{sec:results:performance}
\begin{figure}[t]
    \centering
    \begin{tabular}[]{@{}l@{\hspace{0.2em}}c@{\hspace{0.2em}}c}
    \multicolumn{3}{c}{\includegraphics[width=0.8\textwidth]{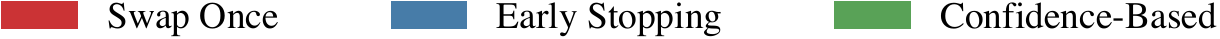}}
    \\
        & Task: Re-rank
        & Task: Alignment
        \\
        \rotatebox[origin=lt]{90}{\hspace{2em} \scriptsize Normalized Accuracy}
        & \includegraphics[width=0.45\textwidth]{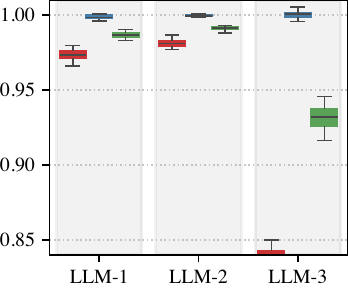}
        & \includegraphics[width=0.45\textwidth]{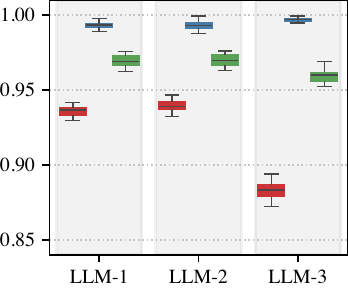}
    \end{tabular}
    \caption{The accuracy of different methods for judgments, normalized by the consensus judgment as the skyline. As both early stopping methods are probabilistic, there is non-zero probability that the accuracy of them exceeds that of the consensus judgment.}
    \label{fig:accuracy}
\end{figure}

\cref{fig:accuracy} presents the main results of our work: a comparison of the accuracy of our early stopping methods normalized by the consensus judgment (which serves as the upper bound in accuracy). 
Additionally, \emph{Swap Once} represents the scenario where each ordering is presented to the LLM only once, serving as the lower bound in accuracy while being the most cost-efficient approach.

\begin{table}[]
\small
    \centering
    \caption{Average number of LLM calls per judgment instance.}
    \label{tab:runs}
\begin{tabular}{l ccc ccc}
\toprule
 & \multicolumn{3}{c}{Alignment} & \multicolumn{3}{c}{Re-rank} \\
 \cmidrule(lr){2-4} \cmidrule(lr){5-7} 
 & {\cellcolor[gray]{0.75} \llmb{}} & {\cellcolor[gray]{0.92} \llmm{}} & { \llmo{}} & {\cellcolor[gray]{0.75} \llmb{}} & {\cellcolor[gray]{0.92} \llmm{}} & { \llmo{}} \\
\midrule
{ Swap Once} & \cellcolor[gray]{0.75} 2 & \cellcolor[gray]{0.92} 2 & 2 
& \cellcolor[gray]{0.75} 2 & \cellcolor[gray]{0.92} 2 & 2 \\
{ Early Stopping} & \cellcolor[gray]{0.75} 3.60 & \cellcolor[gray]{0.92} 3.33 & 9.43 
& \cellcolor[gray]{0.75} 3.58 & \cellcolor[gray]{0.92} 3.11 & 4.83 \\
{ Confidence-Based} & \cellcolor[gray]{0.75} 2.69 & \cellcolor[gray]{0.92} 2.59 & 4.71 
& \cellcolor[gray]{0.75} 2.65 & \cellcolor[gray]{0.92} 2.43 & 2.93 \\
{ Consensus Outcome} & \cellcolor[gray]{0.75} 24 & \cellcolor[gray]{0.92} 24 & 24 
& \cellcolor[gray]{0.75} 24 & \cellcolor[gray]{0.92} 24 & 24 \\
\bottomrule
\end{tabular}
\end{table}

The results show that the original early-stopping method (\cref{sec:judgement:earlystopping}) matches the accuracy of the consensus judgment in all cases ---i.e., reaching a normalized accuracy of \(1\)---, while the confidence-based early stopping (\cref{sec:judgement:confidencebased}) closely follows. Both methods significantly outperform the Swap Once approach in accuracy.
These improvements come from selectively increasing repetitions only for harder instances. Specifically, instances that exhibit PC (\cref{tab:bias-llm}) require just one paired repetition to reach peak accuracy, making all methods as effective as the consensus judgment for these cases. However, for non-PC instances, Swap Once is not enough to achieve a robust judgment, necessitating additional repetitions.

The significance of these results becomes evident when considering the average number of required LLM calls presented in \cref{tab:runs}. 
We observe that the Early Stopping method, despite reaching the same accuracy as the consensus judgment, reduces the average number of LLM calls by more than \( 81\% \). Moreover, the Confidence-Based method further reduces the average number of LLM calls by an average of \( 87\%\) compared to consensus judgment, requiring just slightly more than two calls but achieves significantly higher accuracy than Swap Once.\looseness=-1


\section{Related Work}
Recent studies have investigated position bias and repeat inconsistency, e.g., \cite{shi2024judgingthejudges} \cite{wang2024largelanguagemodels}.
These studies have found that position bias is not random; its magnitude and direction can vary across different tasks and models. 
Notably, the quality gap between solutions significantly influences the extent of this bias.
Similar findings are also reported in other research, e.g., \cite{guo2024biasinlarge} \cite{li2024calibraevalcalibratingprediction} \cite{gu2024asurveyon} \cite{li2024fromgenerationto} \cite{ye2024justiceorprejudice?} \cite{zhu2024starling}.

Several strategies have been proposed to mitigate position bias in LLM-based evaluations. 
A great number of studies propose to shuffle the order of texts before judgment and average the scores, e.g., \cite{li2023generativejudgeevaluatingalignment} \cite{zhu2023judgelmfinetunedlargelanguage} \cite{shi2024judgingthejudges} \cite{sottana-etal-2023-evaluation} \cite{chen2024mllmasajudgeassessingmultimodalllmasajudge}. 
Building on these insights, we introduced an adaptive, instance-specific approach to repetition-based judgment refinement, that, instead of using a fixed number of repetitions, dynamically adjusts the number of repetitions based on instance difficulty, increasing repetitions for harder cases to boost accuracy while reducing overall computational costs.
In contrast to the shuffle and repeat approach, \cite{li2024split} proposes to split the candidates into meaningful segments and merge them back together to reduce the effect of position bias in LLM evaluator.
While effective in its context, this approach is fundamentally different from ours and targets a distinct mitigation mechanism; thus, it is not directly comparable as a baseline to our method.

Another line of research takes a mechanistic approach to mitigate position bias by analyzing LLM internals to identify and adjust the components responsible for such biases \cite{li2024calibraevalcalibratingprediction} \cite{wang2024eliminatingpositionbias} \cite{zheng2024largelanguagemodels}. While these methods aim to provide a universal correction, our findings suggest that a single treatment is suboptimal due to the varying direction and magnitude of position bias across instances. Instead, our approach dynamically adapts to each judgment instance, ensuring a more targeted and effective mitigation strategy.


\section{Conclusion}
In this paper, we explored the impact of position bias in LLMs on judgment consistency and proposed a novel early-stopping method to reduce computational costs while maintaining robust and accurate consensus judgments. 
Our experimental results demonstrate the effectiveness of early stopping in minimizing LLM calls without significantly sacrificing accuracy, making it a promising approach for large-scale applications. 

Looking ahead, there are several avenues for future work. First, further investigations could focus on analyzing the reasons behind the varying direction of position bias within a single dataset. Understanding these underlying factors could lead to more precise treatments for position bias. 
Additionally, extending our early-stopping methods to more complex judgment scenarios, such as multi-item comparisons, could further enhance efficiency without compromising accuracy. 
Lastly, our observation of consistency when the inherent judgment aligns with position bias could be tested for other types of bias.

% ...

\bibliographystyle{splncs04}
\bibliography{biblio/base}


\end{document}